\documentclass[a4paper, 10pt]{article}
\usepackage{latexsym}
\usepackage{amssymb}
\usepackage{algorithmic}
\usepackage{graphicx}
\usepackage{psfrag}
\usepackage{subfigure}

\newcommand{\PROCEDURE}[1]{\item[\textbf{procedure}] #1}
\newcommand{\BEGIN}{\item[\textbf{begin}]{}}
\newcommand{\END}{\item[\textbf{end}]{}}

\newcommand{\ignore}[1]{}
\newcommand{\pjsignore}[1]{}

\newtheorem{theorem}{Theorem}
\newtheorem{proposition}{Proposition}
\newtheorem{lemma}{Lemma}
\newtheorem{corollary}{Corollary}
\newenvironment{proof}{\noindent {\bf \em Proof:}}{\hfill $\Box$ 
\bigskip}

\begin{document}

\title{Islands for SAT}
\author{
    H.\ Fang\thanks{Department of Computer Science, Yale University,  
USA.
    Email:\ {\tt hai.fang@yale.edu}}
    \and Y.\ Kilani\thanks{Department of Computer Science and  
Engineering,
    The Chinese University of Hong Kong, Shatin, N.T., Hong Kong, China.
    Email:\ {\tt $\{$ykilani,jlee$\}$@cse.cuhk.edu.hk}}
    \and J.H.M.\ Lee$^\dagger$ \and P.J.\ Stuckey\thanks{
    NICTA Victoria Laboratory,
    Department of Computer Science and Software Engineering,
    University of Melbourne,
    Parkville 3052, Australia. Email:\ {\tt pjs@cs.mu.oz.au}}}
\maketitle

\begin{abstract}
In this note we introduce the notion of islands for restricting local
search.  We show how we can construct islands for CNF SAT problems,
and how much search space can be eliminated by restricting search to
the island.
\end{abstract}

\section{Background and Definitions} \label{back}

In the following subsections, we give the necessary definitions and
notations for subsequent discussion and presentation.

\subsection{SAT}
A {\em (propositional) variable\/} can take the value of
either 0 (false) or 1 (true).
A {\em literal\/} is either a variable $x$ or its complement $\bar{x}$.
A literal $l$ is {\em true\/} if $l$ assumes the value 1; $l$ is
{\em false\/} otherwise.
A {\em clause\/} is a disjunction of literals, which is true when one of
its literal is true.
A {\em Satisfiability (SAT)\/} problem consists of
a finite set of variables and a finite set of clauses (treated as
conjunction).

A SAT problem is a special case of
a CSP $(Z,D,C)$:\ $Z$ is the set of variables of the
SAT problem, the
domain of each variable is $\{0,1\}$, and $C$ contains all the
clauses, each of which is considered
a constraint in $C$ restricting the values that the variables can take.

Given a CSP $P = (Z,D,C)$.
We use $var(c)$ to denote the set of variables
that occur in constraint $c \in C$.
A \emph{valuation} for variable set $\{x_1, \ldots, x_n\} \subseteq Z$
is a mapping from variables to values
denoted $\{x_1 \mapsto a_1, \ldots, x_n \mapsto a_n\}$
where each $x_i$ is a variable and $a_i \in D_{x_i}$.

A {\em state\/} of $P$ (or $C$) is a valuation for $Z$.
The {\em projection\/} $\pi(s,v)$ of a valuation $s$ on variable set  
$v'$
onto a set of variables $v \subseteq v'$
is defined as
\[ \pi(s,v) = \{ x \mapsto a \,|\, (x \mapsto a \in s) \wedge (x \in  
v)\}. \]
A state $s$ is a {\em solution\/} of a constraint $c$ if $\pi(s,var(c))$
is a set of variable assignments which makes $c$ true.
A state $s$ is a {\em solution\/} of a CSP $(Z,D,C)$ if $s$ is a  
solution
to all constraints in $C$ simultaneously.  In the context of SAT
problems, a solution makes all clauses true simultaneously.

Since we are dealing with SAT problems we will also use an alternate
representation of a state as a set of literals.
A state $\{x_1 \mapsto a_1, \ldots, x_n \mapsto a_n\}$
corresponds to a set of literals $\{ x_j ~|~ a_j = 1\} \cup \{\bar{x} 
_j ~|~
a_j = 0\}$.

Unless stated otherwise, we understand constraints (or clauses) in
a set as always conjuncted.  Therefore, we abuse terminology by
using the phrases ``a conjunction of constraints (or clauses)'' and  
``a set of
constraints (or clauses)'' interchangeably.

\subsection{Local Search}

A local search solver moves from one state to another using a local  
move.
We define the {\em neighbourhood\/} $n(s)$ of a state $s$
to be all the states that are reachable in a single move from state $s$.
The neighbourhood states are meant to represent all the states reachable
in one move, independent of the actual heuristic function used to choose
which state is moved to.

For the purpose of this paper, we assume the
neighbourhood function $n(s)$ returns the states which are
at a Hamming distance of 1 from the starting state $s$.
The Hamming distance between states $s_1$ and $s_2$ is defined as
\[ d_h(s_1,s_2) = |s_1 - (s_1 \cap s_2)| = |s_2 - (s_1 \cap s_2)|. \]
In other words, the Hamming distance measures the number of  
differences in
variable assignment of $s_1$ and $s_2$.
This neighbourhood reflects the usual kind of local move in
SAT solvers, \emph{flipping} one variable.

A {\em local move\/} from state
$s$ is a transition, $s \Rightarrow s'$, from $s$ to $s' \in n(s)$.  A
{\em local search procedure\/} consists of at least the following  
components:
\begin{itemize}
\item a neighbourhood function $n$ for all states;
\item a heuristic function $b$ that determines the ``best'' possible  
local
       move $s \Rightarrow s'$ for the current state $s$; and
\item possibly an optional ``breakout'' procedure to help escape from
       local minima.
\end{itemize}
We note that the notion of noises as appeared in some solvers,
such as WalkSAT, can be incorporated into the
heuristic function $b$.
We also decouple the notion of neighbourhood
from the heuristic function since they are orthogonal to each other,
although they are mixed together in the description of a local move
in GSAT, WalkSAT, and others.

\section{Island Constraints} \label{island}

We introduce the notion of {\em island constraints\/}, the solution
space of which is connected in the following sense.  Central to a local
search algorithm is the definition of the neighbourhood of a state since
each local move can only be made to a state in the neighbourhood of the
current state.  We say that a constraint is an island constraint if we
can move from any state in the constraint's solution space to another
using a sequence of local moves without moving out of the solution  
space.

Let $sol(c)$ denote the set of all solutions to a constraint $c$,
in other words the {\em solution space\/} of $c$.
A constraint $c$ is
an \emph{island constraint} (or simply \emph{island}) if, for any
two states $s_0,s_n \in sol(c)$, there exist states
$s_1,\ldots,s_{n-1} \in sol(c)$ such that $s_i \Rightarrow s_{i+1}$
for all $i \in \{0,\ldots,n-1\}$.
A constraint $c$
with $|sol(c)| \leq 1$ is thus an island by definition.
We call such islands {\em trivial\/}.

Immediately questions about islands arise:
\begin{itemize}
\item When is a constraint an island?
\item Given $n$ islands $c_1,\ldots,c_n$ of different constraint  
types.  When
       is the conjunction $c_1 \wedge \cdots \wedge c_n$ an island,  
if at all?
\end{itemize}

Before embarking on answering these questions,
without loss of generality, we assume from now on that all
clauses are in {\em standard form\/}:
(1) no literals occur more than once in the same clause,
and (2) no literal and its complement occur together in the same clause.
This standard form requirement
is easy to fulfill since we observe that
\[ \cdots \vee l \vee \cdots \vee l \vee \cdots \equiv \cdots \vee l  
\vee \cdots \]
and
\[ \cdots \vee l \vee \cdots \vee \bar{l} \vee \cdots \equiv true \]
for any literal $l$.

\begin{theorem}
Any clause $c$ forms an island.
\end{theorem}
\begin{proof}
Consider two solutions $s_0$ and $s_n$ of $c$.  Then
(treating them as sets of literals) $s_0 \cap c \neq \emptyset$
and $s_n \cap c \neq \emptyset$. Choose
$l_n \in s_n \cap c$. Clearly
$s_1 = s_0 - \{\bar{l}_n\} \cup \{l_n\}$ is also a solution of
$c$, and either equals $s_0$ or is a neighbour.
Now move from $s_1 \Rightarrow^* s_n$ be flipping any variable
different from that in $l_n$. Clearly each state in this sequence
is a solution becuase is contains $l_n$.
\end{proof}

\section{Non-Conflicting Clause Set} \label{nonConfl}

We give a first sufficient condition for
when a set $C$ of clauses
results in an island.
We note that any solution to a clause must
contain at least one assignment of the form $l/1$.
The idea is to disallow
the simultaneous occurrences of $l$ and $\bar{l}$ in $C$.
The intuition of this restriction is as follows.
Suppose literal $l$ occurs in clause $c_i$ and $\bar{l}$ occurs
in $c_j$.  Suppose $l$ is 0.  During the course of the local moves,  
it might
be necessary to set $l$ to 1.  However, if $\bar{l}$ is the only literal
in $c_j$ assuming the value 1, resetting $\bar{l}$ falsifies $c_j$,  
moving
the trajectory out of $sol(C)$.

Let $lit(c)$ denote the set of all literals of a clause $c$.  A set $C 
$ of
clauses is {\em non-conflicting\/} if there does not exist a variable
$x$ such that $x,\bar{x} \in \bigcup \{lit(c)\,|\,c \in C\}$.

\begin{theorem} \label{nonConflTh}
A non-conflicting set $C$ of clauses forms an island.
\end{theorem}
\begin{proof}
Consequence of Theorem~\ref{primalIsland} proved in the following  
section.
\end{proof}

\section{Primal Non-Conflicting Clause Set} \label{primal}

The requirement of the non-conflicting property on all variables is
too stringent.
It suffices to impose this restriction on only a subset of variables,
in particular, only one variable from each clause.

Without loss of generality,
we impose an arbitrary total ordering $<$ on the
variables in a SAT problem.
With such a total ordering, it makes sense
to talk about the {\em least\/} variable among a set of variables.
We say that $l$ is the {\em $<$-primal literal\/},
denoted by $p_<(c)$,
of a clause $c$ if
$var(l)$ is the least among all variables in $var(c)$ using
the $<$ ordering.

Given a set of clauses $C$ and a variable ordering $<$.
The {\em $<$-primal literal set\/} of $C$, $pLit_<(C)$, is the
set of all $<$-primal literals of the clauses in $C$.
In other words,
\[ pLit_<(C) = \{p_<(c) \,|\, c \in C\}. \]
$C$ is {\em $<$-primal non-conflicting\/} if
there does not exist a variable
$x$ such that $x,\bar{x} \in pLit_<(C)$.

\begin{lemma} \label{exist}
Given a $<$-primal non-conflicting set $C$ of clauses with variable
ordering $<$ any state $s \supseteq pLit_<(C)$ is a solution of $C$.
\end{lemma}
\begin{proof}
Since every clause in $C$ contains a literal from $pLit_<(C)$, the  
variable
assignments in $s$ make at least one literal in each clause true.
\end{proof}

Lemma~\ref{exist} gives a method to find a solution of $C$.  This  
solution
consists of any assignments that makes the literals in $pLit_<(C)$ true.
The
assignments for variables not in $pLit_<(C)$ can be arbitrary.  For  
example,
if $C$ has variables $\{x_1,\ldots,x_5\}$ and
$pLit_<(C) = \{\bar{x_2},x_4,\bar{x_5}\}$, then
\[ \{x_1/1,x_2/0,x_3/1,x_4/1,x_5/0\} \]
is a solution of $C$.  Note that the
assignments for variables $x_1$ and $x_3$ can be arbitrary since they
are not in $pLit_<(C)$.

\begin{theorem} \label{primalIsland}
A $<$-primal non-conflicting set $C$ of clauses forms an island.
\end{theorem}
\begin{proof}
Given any solutions $s$ of $C$
we construct a path of moves (remaining as solutions of $C$)
from $s$ to $\hat{s}$ where $\hat{s} \supseteq pLit_<(C)$.
Clearly we can move from any solution $\hat{s}$ to another
$\hat{s}'$ where $\hat{s}' \supseteq pLit_<(C)$ simply by modifying
literals not in $pLit_<(C)$.
Hence we have a path from any solution to any other.

Suppose $pLit_<(C) \not\subseteq s$.
There must exist a least variable
$x$ such that the either $\bar{x} \in s$ and $x \in pLit_<(C)$
or $x \in s$ and $\bar{x} \in pLit_<(C)$.
Let $l$ be the literal in $s$ containing $x$.
Define $s' = s - \{l\} \cup \{\bar{l}\}$.

Consider each clause $c \in C$, we show that $s'$ is a solution of  
each $c$.
\begin{itemize}
\item $p_<(c) = \bar{l}$:
Clearly $s'$ is a solution of $c$.
\item $p_<(c) = l$: Contradiction since $\bar{l} \in pLit_<(C)$ and $C$
is $<$-primal non-conflicting.  Hence this case cannot occur.
\item $p_<(c)$ involves variable $x' < x$: By the choice of $x$, we have
that $p_<(c) \in s$ and hence also in $s'$. Thus $s'$ is a solution  
of $c$.
\item $p_<(c)$ involves variable $x' > x$: Clearly the variable
$x$ does not occur in $c$ (otherwise it would give the primal literal).
Since the only difference between $s'$ and $s$ is on $x$, clearly $s'$
remains a solution of $c$.
\end{itemize}

Since the number of literals in $s' \cap pLit_<(C)$ is one more than
in $s \cap pLit_<(C)$, this process eventually terminates in a solution
$\hat{s} \supseteq pLit_<(C)$.
\end{proof}

Note that that the total ordering on variables
is entirely arbitrary.  It gives us a consistent way of picking a
primal literal for each clause $c$, and thus moving from any solution  
to any
other, through the primal literal set.

A direct consequence of Theorem~\ref{primalIsland} is its converse,  
stated
as follows.
\begin{corollary} \label{primalIslandConv}
If a set $C$ of clauses is satisfiable but not an island, then
there exists no ordering $<$ such that $C$ is $<$-primal non- 
conflicting.
\end{corollary}

Consider an island $C$ formed from a set of constraints.
If every subset of $C$ is also an island, we say that $C$ is
{\em compositional\/}.
\begin{proposition} \label{primalToCompo}
Given any total ordering $<$ on variables.
Islands formed from $<$-primal non-conflicting sets of clauses are
compositional.
\end{proposition}
\begin{proof}
Suppose the set $C$ of clauses is $<$-primal non-conflicting.
We observe that every subset of $C$ is also $<$-primal non-conflicting.
Therefore, every subset of $C$ is an island.
\end{proof}

We shall see later that compositionality
is important for the dynamic
version of the Island Confinement Method.
The
converse of Proposition~\ref{primalToCompo} does not
hold.  Consider the simple island
\[ C = (x_1 \vee x_2 \vee \bar{x}_3) \wedge
        (\bar{x}_1 \vee \bar{x}_2 \vee x_3) \]
which is compositional since any individual clause forms an island.
We can also easily verify that there exists no ordering $<$ that makes
$C$ $<$-primal non-conflicting.  It is because the two clauses $c_1$
and $c_2$ in $C$ are ``mirror images'' of each other in the sense that
for every literal $l$ in $c_1$, $\bar{l}$ is in $c_2$, and {\em vice  
versa\/}.
Thus, no matter what the ordering $<$ is, we would have both $l$ and $ 
\bar{l}$
in the $<$-primal literal set.
This means that the $<$-primal non-conflicting
property is only a {\em sufficient\/} but {\em not\/} a
necessary condition for
compositional islands or even just island.  The search for a
more exact characterization of islands continues.

On the other hand, we show in the next two sections that
$<$-primal non-conflicting sets cover a large class, although not all,
of islands, and are useful in practice.  Given a SAT problem
$C$.  We give a greedy algorithm to compute a $<$-primal non-conflicting
subset of $C$.  Our results show that this subset covers over 80\% of  
the
clauses on average using 11 benchmarks from the DIMACS archive.

\section{A Greedy Algorithm} \label{greedy}

Figure~\ref{greedyAlg} gives a simple greedy algorithm, \textsf{islandExtr},
\begin{figure}
\begin{center}
\parbox{0.65\textwidth}{
\begin{algorithmic}
\PROCEDURE{\textsf{islandExtr}($C$:in,$L$:out,$Q$:out)}
\BEGIN
    \STATE $L \leftarrow []$;
    \STATE $Q \leftarrow \emptyset$;
    \WHILE{$C \not= \emptyset$}
       \STATE pick the ``best'' literal $l$ in $C$;
       \STATE $L \leftarrow L $$+$$+$$ [l]$;
       \STATE $Q \leftarrow Q \cup \{\mbox{all clauses in $C$  
containing only $l$}\}$;
       \STATE $C \leftarrow C - \{\mbox{all clauses in $C$ containing  
either $l$ or $\bar{l}$}\}$;
    \ENDWHILE
\END
\end{algorithmic}
}
\end{center}
\caption{The \textsf{islandExtr} greedy algorithm} \label{greedyAlg}
\end{figure}
for extracting a
$<$-primal non-conflicting subset of clauses from an arbitrary set of
clauses.  The input to the algorithm is a set of clauses, and the output
is a $<$-primal non-conflicting set $Q \subseteq C$ of clauses plus the
the $<$-primal literal set $L$ (stored as a list) of $Q$.
The ordering of the literals in the list $L$ induces a variable  
ordering $<$,
which is divided into two parts.
The ordering of the variables in $L$ follows the same ordering of their
corresponding literals in $L$.  The ordering among variables not in $L 
$ can be
arbitrary but they must all be {\em greater than\/} variables in $L$.
It should be noted that $L$, which is essentially a sequenced version
of $pLit_<(Q)$, gives
also a solution to the output island $Q$ using Lemma~\ref{exist}.

The \textsf{islandExtr}
algorithm works as follows.
Initially $L$ and $Q$ are
empty, ready to accumulate results to be collected.  While there are
still clauses from $C$, the algorithm tries to find
the ``best'' literal $l$ from $C$.
We defer our discussion of the notion of ``best'' to the
next paragraph, in order not to break the flow of the description of the
algorithm.
This ``best'' literal will be the $<$-prime literal in all
clauses containing $l$ in $C$, which will be added to $Q$ to become part
of the $<$-primal non-conflicting set that we are computing.
That is why $l$ is appended to $L$.
The $+$$+$ operator stands for list concentenation.
Now clauses containing $l$ can be removed from $C$ since they are  
already
in $Q$.  Clauses containing $\bar{l}$ must also be removed since
$\bar{l}$ is the prime literal of these clauses, which can never qualify
to be added to $Q$.  This process is repeated until $C$ becomes empty.

The objective of the \textsf{islandExtr}
algorithm is to collect as many clauses
from $C$ as possible for $Q$, which is determined directly by the  
choice of
$l$ in each step of the loop.  We encode greedy heuristics in the  
selection
of the ``best'' literal.  One naive approach is to select the literal
$l$ that occurs in the most number of clauses in $C$.  What could go  
wrong,
however, is that a large number of clauses containing $\bar{l}$ might  
also
be removed as a result of this selection.  Therefore, the greedy  
heuristic
should strike a careful balance between the number of clauses containing
$l$ and those containing $\bar{l}$.  The idea is that the benefit gained
from selecting $l$ should outgrow the penalty for removing clauses  
containing
$\bar{l}$.  Some possibilities are to choose the literal $l$ with
the maximum of the following expressions:
\begin{itemize}
\item $- \#(\bar{l})$,
\item $\#(l) - \#(\bar{l})$,
\item $\#(l) / \#(\bar{l})$, and
\item $\#(l) / (\#(l)+\#(\bar{l}))$,
\end{itemize}
where $\#(l)$ denotes the number of clauses containing $l$ as a literal
in $C$.
Note that the second and the third expressions are equivalent
since
\[\#(l_1)\#(l_2)+\#(l_1)\#(\bar{l_2}) \geq \#(l_2)\#(l_1)+\#(l_2)\# 
(\bar{l_1})\]
implies
\[\#(l_1)\#(\bar{l_2}) \geq \#(l_2)\#(\bar{l_1}).\]
Different expressions above give a different metric to measure
the ``efficiency'' of $l$ over $\bar{l}$ as compared to other literals
in $C$.  More complex heuristics can be devised, but we should bear in
mind that greedy algorithms are supposed to be simple and efficient.

Table~\ref{hard} gives the result of applying the
\textsf{islandExtr} algorithm
\begin{table}
\begin{center}
\begin{tabular}{|l|c|c|c|c|}
\hline
& $|C|$ & $|Q|$ & $|var(C)|$ & $|n(L)|$ \\
\hline
aim\_100\_1\_6 & 160 & 150 (93.8\%) & 100 & 38 (38\%) \\ \hline
hanoi4 & 4934 & 4065 (82.4\%) & 718 & 197 (27.4\%) \\ \hline
f600 & 2550 & 2134 (83.7\%) & 600 & 183 (30.5\%) \\ \hline
f2000 & 8500 & 7072 (83.2\%) & 2000 & 624 (31.2\%) \\ \hline
\end{tabular}
\end{center}
\caption{Greedy Algorithm on Hard DIMACS Problems} \label{hard}
\end{table}
to four hard problems in the DIMACS archive.
The expression ``$\#(l) / \#(\bar{l})$'' is used to select the best  
variable.
These are large problems containing 100 to 2000 variables.
The first column contains the problem names.
The second column gives the number of clauses.  The third gives
the number of clauses of the extracted island and its associated  
percentage.
The fourth column gives the total number of variables.  The last column,
denoted by $|n(L)|$, gives the size of the neighbourhood of the initial
solution (obtained from $L$ using Lemma~\ref{exist}) restricted
to only states on the islands.  For example, each state in
``aim\_100\_1\_6'' (which has 100 variables) has 100 neighbouring  
states.
If we restrict our attention to only states in the island extracted,
the initial solution has only 38 neighbouring states.

To further demonstrate the benefits of identifying islands in a SAT
problem, we performed the same experiment on a set of small problems,
also from the DIMACS archive.  Each of these problems contains 20  
variables
and 91 clauses.  Therefore, the size of the entire search space of each
problem is $2^{20} = 1,048,576$ in terms of the number of states.  We  
choose
small problems so that we can use a complete
search algorithm to find the size of the search space of the extracted
islands and the number of solutions, which are reported in the third and
fourth columns of Table~\ref{easy}.  The number and percentage of
\begin{table}
\begin{center}
\begin{tabular}{|c|c|c|c|c|}
\hline
& $|Q|$ & $|Space(Q)|$ & $2^{20}/|Space(Q)|$ & $|sol(C)|$ \\ \hline
uf20-01 & 72 (79.1\%) & 1300  &    807      & 8 \\ \hline
uf20-99 & 74 (81.3\%) & 1175  &    892      & 8 \\ \hline
uf20-300 & 78 (85.7\%) & 537  &   1952      & 8 \\ \hline
uf20-500 & 72 (79.1\%) & 879  &     119     & 3 \\ \hline
uf20-800 & 72 (79.1\%) & 683  &      1535   & 8 \\ \hline
uf20-999 & 75 (94.9\%) & 416  &   2521      & 23 \\ \hline
uf20-1000 & 70 (76.9\%) & 1070&    980      & 1 \\ \hline
\end{tabular}
\end{center}
\caption{Greedy Algorithm on Easy DIMACS Problems} \label{easy}
\end{table}
clauses of the extracted islands are reported in the second column of
the table.

Of the eleven benchmarks that we tried, the islands contain on average
over 80\% of the total number of clauses of the corresponding problems.
Experiments on the smaller problems also demonstrate an actual reduction
of {\em three orders of magnitude\/}
in the search space of the islands over
that of the original problems.
Of course the question remains whether the smaller search space actually
helps the local search algorithm.

\end{document}